\documentclass{article}
\usepackage[round]{natbib}
\usepackage[preprint]{neurips_2024}
\usepackage[utf8]{inputenc} 
\usepackage[T1]{fontenc}    
\usepackage{hyperref}       
\usepackage{url}            
\usepackage{booktabs}       
\usepackage{amsfonts}       
\usepackage{nicefrac}       
\usepackage{microtype}      

\usepackage{lipsum}
\usepackage{fancyhdr}       
\usepackage{graphicx}       
\graphicspath{{media/}}     
\usepackage{amsmath}
\usepackage{float}
\usepackage{algorithm}
\usepackage{algpseudocode}
\usepackage{graphicx}
\usepackage{subcaption} 
\usepackage[table]{xcolor}
\usepackage{multirow}
\usepackage[most]{tcolorbox} 

\title{Semantic Soft Bootstrapping: Long Context Reasoning in LLMs without Reinforcement Learning}

\author{
  Purbesh Mitra \\
  University of Maryland \\
  \texttt{pmitra@umd.edu} \\
   \And
  Sennur Ulukus \\
  University of Maryland \\
  \texttt{ulukus@umd.edu} \\
}

\begin{document}
\maketitle

\begin{abstract}
Long context reasoning in large language models (LLMs) has demonstrated enhancement of their cognitive capabilities via chain-of-thought (CoT) inference. Training such models is usually done via reinforcement learning with verifiable rewards (RLVR) in reasoning based problems, like math and programming. However, RLVR is limited by several bottlenecks, such as, lack of dense reward, and inadequate sample efficiency. As a result, it requires significant compute resources in post-training phase. To overcome these limitations, in this work, we propose \textbf{Semantic Soft Bootstrapping (SSB)}, a self-distillation technique, in which the same base language model plays the role of both teacher and student, but receives different semantic contexts about the correctness of its outcome at training time. The model is first prompted with a math problem and several rollouts are generated. From them, the correct and most common incorrect response are filtered, and then provided to the model in context to produce a more robust, step-by-step explanation with a verified final answer. This pipeline automatically curates a paired teacher-student training set from raw problem-answer data, without any human intervention. This generation process also produces a sequence of logits, which is what the student model tries to match in the training phase just from the bare question alone. We set up an experiment to train the open-source model Qwen2.5-3B-Instruct on GSM8K dataset via parameter-efficient fine-tuning. We then tested its accuracy on MATH500, and AIME2024 benchmarks. Our experiments show a jump of 10.6\%, and 10\% improvements in accuracy, respectively, over group relative policy optimization (GRPO), which is a commonly used RLVR algorithm. Our code is available at \url{https://github.com/purbeshmitra/semantic-soft-bootstrapping}, and the model, curated dataset is available at \url{https://huggingface.co/purbeshmitra/semantic-soft-bootstrapping}.
\end{abstract}

\section{Introduction}
Recently, long context reasoning in large language models (LLMs) have shown enormous progress in accelerating different aspects of scientific and engineering tasks. The most notable examples are the rise of LLM assisted programming~\citep{watanabe2025use, ge2025survey}, achieving top human level performance in math examinations like IMO gold medal~\citep{huang2025gemini}, human level performance at ARC-AGI pattern recognition test~\citep{chollet2024arc}, and so on. Very recently, there have been several scientific breakthroughs facilitated by LLM based systems~\citep{bubeck2025early, rizvi2025scaling}. Behind all such reasoning LLMs, is the post-training paradigm of reinforcement learning with verifiable rewards (RLVR). Starting from the open-source implementation of group relative policy of optimization (GRPO)~\citep{shao2024deepseekmath} in Deepseek-R1 model~\citep{guo2025deepseek}, numerous such optimizing techniques are being used for RL training of LLMs~\cite{liu2025understanding, yu2025dapo, zheng2025group}. The basic idea of RLVR is letting the LLM generate rollouts on a multitude of reasoning questions with verifiable answers. The reinforcement learning algorithm then rewards the LLM weights more, which has the higher probability of generating correct answers to the questions. As a result, the chain-of-thought (CoT) of the LLM becomes longer and sophisticated reasoning pattern emerges in the thought traces.

Even though, RLVR has delivered quite a lot of success in the advancement of LLM reasoning capabilities, there are several bottlenecks and limitations of such approaches. For example, in the outcome-based reward formulation, the reward in RLVR is not dense, i.e., each trajectory of an LLM response gets a coarse reward at the end of the answer. On top of that, the RL algorithm takes an average of those rewards for estimating the success probability of the LLM. This leads to the LLM not being able to fully grasp the underlying reasoning in different tasks. If an LLM generates two responses, one of which is completely wrong logic from the beginning, and the other one has a ``silly mistake'' at the very end of the response, both of them are assigned the same low reward. On the other hand, if a response has a wrong reasoning in it, but still it manages to somehow generate the correct answer, it gets incentivized to follow such reasoning again. To combat this, if we employ some kind of process supervision of the LLM reasoning traces, it is difficult to assign partial rewards to the logical steps without any possibility of reward hacking. Furthermore, in some recent studies~\citep{yue2025does, wu2025invisible}, it has been shown that the fundamental reasoning capabilities of the LLM do not increase with RLVR. Rather, the base model already has some reasoning capabilities, which are then amplified into the pass@1 accuracy of the RL trained model from pass@k, thus, making its capabilities even narrower.

These issues are largely underexplored topics, and only a handful of ideas have been proposed as alternatives. \cite{agrawal2025gepa} introduce a reflective prompt optimizer that combines natural-language self-analysis with multi-objective evolutionary search over prompts. This method outperforms GRPO and prior prompt optimizers while using up to 35 times fewer rollouts across diverse reasoning and tool-use tasks. \cite{karan2025reasoning} show that a training-, dataset-, and verifier-free iterative sampling algorithm targeting a sharpened power distribution defined by base-model likelihoods can elicit reasoning capabilities from base models that can match GRPO post-trained models on benchmarks, such as, MATH500, HumanEval, GPQA, and AlpacaEval, while avoiding the diversity-collapse typical of RLVR post-training. Feedback Descent by \cite{lee2025feedback} is an inference-time framework that optimizes text artifacts, such as, prompts, code using pairwise comparisons augmented with rich textual feedback instead of scalar rewards, thus, treating feedback as a gradient-like signal in text space. Prior self-training work on LLM reasoning, such as STaR~\citep{zelikman2022star}, which iteratively fine-tunes on self-generated rationales that lead to correct answers, Think–Prune–Train-Improve~\citep{costello2025think}, which repeatedly fine-tunes on pruned sets of correct CoT traces, and BOLT~\citep{pang2025bolt}, which bootstraps long chains-of-thought via supervised finetuning on self-synthesized LongCoT data and online RL phase, all rely on standard cross-entropy loss for next-token prediction over textual CoT sequences, often from a teacher model.

All the approaches so far do not take advantage of the in-context learning~\citep{dong2024survey} ability of the LLM itself. RLVR training does the exploration via different rollouts, but incentivizes reward based on the average of individual rollouts. As a result, the LLM does not receive a clear signal about the accuracy of the produced trajectories. Our approach considers using different sample responses by presenting them to the LLM in the context of correctness in turn making better responses. This is similar to the well-studied idea of mixture-of-agents~\citep{wang2024mixture}. Another important aspect is the issue of scalability, which boils down to the idea of bitter lesson~\citep{sutton2019bitter}, i.e., simple algorithms that take advantage of computation outperform other complicated methods. This is one of the reasons why the self-supervised learning (SSL) method~\citep{balestriero2023cookbook} in pre-training scales so well, since simply minimizing the loss function for next-token prediction task provides signal in each token across the whole generation trajectory, thus, scaling with compute power. For this reason, we believe it is necessary to find a compromise between the exploration phase of RL and the signal exploitation in SSL. To the best of our knowledge, there has not been any work that combines RL and SSL in such a manner.

To that regard, we introduce Semantic Soft Bootstrapping (SSB), a self-distillation method that teaches a language model to solve reasoning problems without hints by leveraging its own hinted reasoning as a teacher. We start from a supervised dataset of problems with ground-truth final answers, and first query a base model on the question alone to generate multiple rollouts. These are then partitioned into correct and incorrect attempts using the boxed final answer format. From each problem, we then construct a prompt that includes the original problem, one representative correct solution trace, and one trace with the most common incorrect final answer. The same base model, now used as a teacher under a robust ``refine and explain'' system prompt, produces a single, detailed, corrected solution that passes the final answer check. This can be interpreted as an \emph{in-context contrastive learning} from negative pairs, as opposed to the traditional contrastive learning~\cite{le2020contrastive}. This process yields paired problem-solution data points without human effort. We also extract the teacher model's token-level logits for the answer portion only, and store them as soft labels. During training, a student model (same base model with LoRA adapters) is given only the question, and is optimized to match the teacher’s token distribution on the answer tokens via a KL-based distillation loss, optionally combined with cross-entropy, without any reward model or policy gradient. As a result, the student learns to reproduce robust, step-by-step reasoning and correct final answers from hint-free prompts, while all bootstrapping signal is provided through logit-level self-supervision; without an explicit reinforcement learning algorithm. Additionally, since the student model tries to match verifiably correct response trajectories via matching logits, this method eliminates any chances of reward hacking, which is a considerable issue in RL training. 

Furthermore, since the model is not trained on hard token prediction, but instead minimizing a KL loss with different in-context prompts, we are essentially nudging the model output probability minimally towards the distribution of correct responses. This strictly follows from the fact that KL divergence quantifies the minimum shift of an estimated (student's) probability distribution, to the true (teacher's) probability distribution, hence, retaining the model's generalization capability while training. Since we are not letting the model train on its own generated response, this avoids a collapse in performance of the model~\cite{shumailov2024ai}. Our experiments show that SSB outperforms GRPO in MATH500, and AIME2024 benchmarks by a margin of 10.6\%, and 10\%, respectively. This training was done on a curated set of 256 samples from the GSM8K question-final-answer dataset.

Our contributions are summarized as follows:

\begin{itemize}
    \item We propose Semantic Soft Bootstrapping (SSB), an RL-free self-distillation framework that converts a single base LLM into both teacher and student by exposing them to different views of the same problem (hinted vs. hint-free).
    \item We introduce a semantic refinement stage where the teacher is prompted with both correct and incorrect student-like rollouts and must synthesize a single, correct, robust explanation.
    \item We construct a paired teacher–student dataset in which the teacher receives rich hinted context while the student sees only the raw question, enabling learning to “solve without hints” from self-generated supervision.
    \item We perform logit extraction on the teacher’s robust solutions and design a custom data pipeline that distills only the answer tokens, aligning student predictions with teacher soft labels at the level of token distributions.
    \item SSB shows a gain of 10.6\%, and 10\% improvements in accuracy in MATH500, and AIME2024 benchmarks, respectively, over GRPO, while being trained on a set of 256 curated samples from the GSM8K question-final-answer dataset.
\end{itemize}

\section{Background and Related Works}
DeepSeek-R1~\citep{guo2025deepseek} proposed a pure RLVR approach to improve the reasoning capabilities of LLMs, introducing DeepSeek-R1-Zero and DeepSeek-R1 trained with GRPO~\citep{shao2024deepseekmath} on reasoning questions. By repeatedly sampling chains-of-thought and rewarding correct or high-quality reasoning trajectories, DeepSeek-R1 shows that long, structured reasoning patterns can emerge from an RL signal alone, and that RL can substantially boost pass@k on math and programming tasks. 

Distilling knowledge in a neural network~\citep{hinton2015distilling} is the classical formulation of knowledge distillation, where a large teacher model is compressed into a smaller student model by training the student to match the teacher’s output distribution. The key idea is that soft targets at elevated temperature carry rich information about class similarities, allowing the student to inherit the teacher’s generalization behavior. This approach was shown to improve performance on vision and speech recognition tasks while reducing model size and inference cost. 

On-policy distillation of language models~\citep{agarwal2024policy, lu2025onpolicydistillation} generalizes knowledge distillation to autoregressive sequence models by emphasizing on-policy training. Instead of distilling only on teacher-generated sequences, the student is trained on its own sampled outputs, while the teacher provides targets, which mitigates distribution shift between training and deployment. Their generalized knowledge distillation (GKD) framework optimizes alternative divergences such as reverse KL and integrates seamlessly with RL fine-tuning, demonstrating gains for summarization, translation, and arithmetic reasoning and showing that student-generated mistakes can be turned into a powerful learning signal. Our semantic bootstrapping method is conceptually related in that it also exploits model-generated mistakes, feeding incorrect rollouts into a hinted teacher prompt, but differs in two important ways: We keep the process strictly RL-free (no on-policy sampling during training and no reward optimization), and we use the teacher to synthesize a single corrected explanation from correct and incorrect traces, which is then distilled via stored teacher logits into a student that only ever sees the raw, hint-free questions.

\begin{figure}
    \centering
    \includegraphics[width=\linewidth]{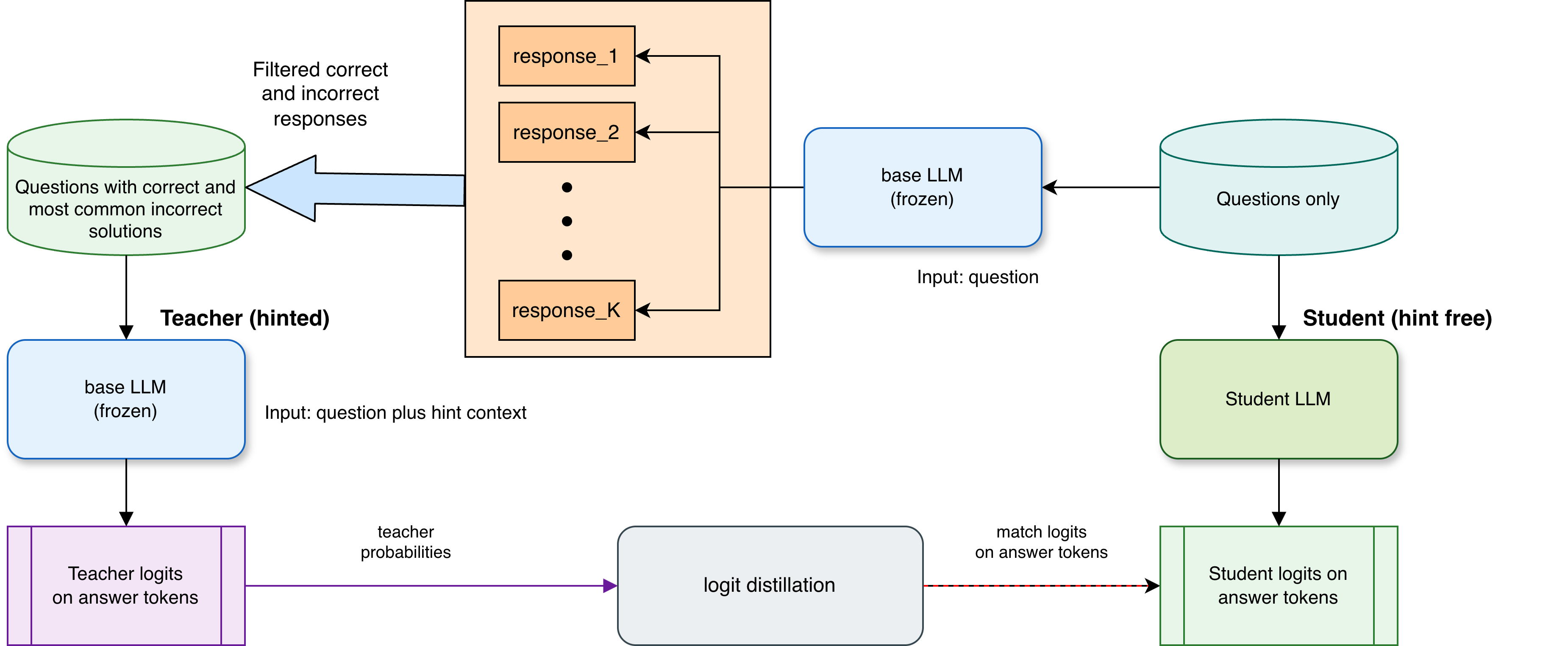}
    \caption{Logit generation and distillation pipeline of SSB algorithm.}
    \label{fig: ssb_method}
\end{figure}

\section{Methodology}

First, we consider a supervised dataset of math problems,
\begin{align}
    \mathcal{D} = \{(q_i, a_i)\}_{i=1}^N,
\end{align}
where $q_i$ is a problem statement and $a_i$ is a discrete final answer (e.g., extracted from \verb|\boxed{...}| in a structured solution). Let $f_\theta$ denote a pre-trained autoregressive language model with parameters $\theta$. It takes system\_prompt, and query as inputs, with a specified temperature, and generates response samples from the distribution $f_\theta$(system\_prompt, query; temperature). The goal is to train $f_{\theta}$ such that, given only the question $q_i$, it produces an answer distribution that matches the teacher's distribution conditioned on a richer, hinted context. The whole methodology is illustrated in Fig.~\ref{fig: ssb_method}.

\subsection{Logit Generation via Multi-Rollout Self-Correction}

\subsubsection{Generating Candidate Solutions}

For each problem $q_i$, we first run the base model multiple times under a simple ``expert tutor'' instruction. Let $\text{sys}_{\mathrm{inf}}$ denote this system message instructing step-by-step reasoning and \verb|\boxed{}| formatting, as following.
\begin{tcolorbox}[colback=gray!10!white,
                  colframe=gray!80!black,
                  arc=2mm, 
                  boxrule=0.5pt, 
                  left=1mm, right=1mm, top=1mm, bottom=1mm]
\texttt{``**Role:** You are an expert math tutor. When you are given a problem to solve, you provide detailed step-by-step reasoning in your solution. Your response is clear, precise, and unambiguous. You do not skip any step. You put your final answer within \textbackslash{boxed}\{\} at the end of your response.''}
\end{tcolorbox}

We perform $K$ stochastic rollouts with sampling temperature $T_{\text{roll}}$:
\begin{align}
    \{r_{k}\}_{k=1}^K \sim f_\theta(\text{sys}_{\mathrm{inf}}, q_i; T_{\text{roll}}),
\end{align}
where each $r_{k}$ is a full solution text. We parse the last \verb|\boxed{...}| expression in each solution to obtain a predicted answer $\tilde{y}_{i,k}$. We then partition the rollouts into two sets, a set of responses which contain the correct final answer, and another set of responses which do not:
\begin{align}
    \mathcal{R}^{\mathrm{correct}} &= \{r_k \mid (r_k, \hat{a}_k) \in \mathcal{R},\ \hat{a}_k = a\}, \quad \text{(correct set)},\\
    \mathcal{R}^{\mathrm{wrong}} &= \{r_k \mid (r_k, \hat{a}_k) \in \mathcal{R},\ \hat{a}_k \neq a\}, \quad \text{(incorrect set)}.
\end{align}
Note that we are only considering the boxed final answer for this evaluation. Even if a response contains the correct answer, but does not enclose it within boxed format, we discard that as incorrect. This is for maintaining a strict format in the response. Instead of assigning format reward like RL, we are essentially performing a kind of rejection sampling for formatting. If either set is empty, i.e., no mix of correct and incorrect attempts are found, we discard $q_i$ from SSB training.

\subsubsection{Selecting Representative Correct and Incorrect Solutions}

From $\mathcal{R}^{\mathrm{wrong}}$, we randomly choose $r^{\mathrm{wrong}}$ as one of the responses with the most common incorrect answer $\hat{a}^{\mathrm{wrong}}$, if such a clear majority exists; otherwise, we sample one incorrect answer at random. We choose the incorrect response like this because, this most common incorrect answer is a representative of the model's inability to grasp the underlying reasoning. We also choose $r^{\mathrm{corr}}$ from $\mathcal{R}^{\mathrm{correct}}$ randomly as the representative of correct solutions. Together, these two form a contrast between a valid solution path and a representative misconception, respectively.

\subsubsection{Teacher Refinement Under Hinted Context}

We now construct a user prompt $u_{\mathrm{SSB}}$ that explicitly shows the model both solutions and asks it to synthesize a robust explanation. The prompt is as following:

\begin{tcolorbox}[colback=gray!10!white,
                  colframe=gray!80!black,
                  arc=2mm, 
                  boxrule=0.5pt, 
                  left=1mm, right=1mm, top=1mm, bottom=1mm]
\texttt{``We have two sample responses from students for a math problem for you to observe. One of the responses is correct, and the other is incorrect. The students were asked to put the final answer inside \textbackslash{boxed}\{\} in their responses. Only this final answer was checked by an automatic evaluator. The following is the problem:\\
-------- \\
{question}\\
-------- \\
\\
This is the correct response from a student:\\
-------- \\
{chosen\_correct\_generation}\\
-------- \\
\\
This is an attempt by another student which was labelled as incorrect by auto-evaluator:\\
-------- \\
{rejected\_generation}\\
-------- \\
\\
Write a coherent, step-by-step derivation of the solution. Do not skip any step in your response, and make it as detailed as possible. This should be a standalone solution of the given problem since this solution will be used by the students for studying and learning. Make the solution robust by cautioning about potential errors or wrong chain of reasoning. However, do not mention in your response that you were provided attempted responses by the students. There should not be a slightest mention or hint that you are actually refining from correct or incorrect responses written by students.\\
\\
Conclude the entire response with the final answer, enclosed in \textbackslash{boxed}\{\}, at the very end. Make sure your enclosed final answer exactly matches the enclosed final answer in the given correct response of the student.''}
\end{tcolorbox}

Conditioned on this hinted context, the model generates a refined solution:
\begin{align}
    \tilde{r}_i \sim f_\theta(\text{sys}_{\mathrm{inf}}, u_{\mathrm{SSB}}; T_{\text{roll}}).
\end{align}
We again parse the final \verb|\boxed{...}| answer from $\tilde{r}_i$ and accept this teacher solution only if it matches the ground-truth $a_i$. For each accepted example, we store:
\begin{itemize}
    \item \textbf{Teacher example} (hinted view): \big[system: $\text{sys}_{\mathrm{inf}}$; user: problem $q_i$ + correct solution $r_i^{\text{corr}}$ + incorrect solution $r_i^{\text{wrong}}$; assistant: refined solution $\tilde{r}_i$\big] in $\mathcal{T}$.
    \item \textbf{Student example} (minimal view): \big[system: $\text{sys}_{\mathrm{inf}}$; user: problem $q_i$; assistant: refined solution $\tilde{r}_i$\big] in $\mathcal{S}$.
\end{itemize}
Both conversations are associated with the same $q_i$ and the same final answer sequence, but the teacher has access to much richer semantic context during generation.

\subsubsection{Precomputing Teacher Logits}
To avoid recomputing teacher logits during training, we precompute and store the logits over the answer sequence. For each teacher example $m_i$ in $\mathcal{T}$, we combine the system and user messages to get a chat template $m_i^{\mathrm{teacher}}$ for the model and then compute the logits sequentially by adding tokens from $\tilde{r}_i$ one by one. We denote the teacher logit corresponding to the $j$th token of $\tilde{r}_i$ as $\ell_i^j = f_\theta(m_i[0:|m_i^{\mathrm{teacher}}|+j])$. We then store the whole teacher logit sequence $\ell_i = \big[\ell_i^1, \ell_i^2, \ldots, \ell_i^{|\tilde{r}_i|}\big]$ in the disk memory.

\subsection{Logit-Level Distillation}
For distilling the generated logits into the model, we first let the model generate its own logits sequentially and match them with teacher logits.

\subsubsection{Knowledge Distillation Objective}
First, for each student example $m_i^{\mathrm{student}} \in \mathcal{S}$, we similarly prepare a chat template $m_i^{\mathrm{student}}$ for the model and then compute the student logits sequentially by adding tokens from $\tilde{r}_i$ one by one. The student logit corresponding to the appended $j$th token of $\tilde{r}_i$ is denoted as $\hat{\ell}_i^j = f_\theta(s_i[0:|m_i^{\mathrm{student}}|+j])$. The generated final student logit sequence is $\hat{\ell}_i = \big[\hat{\ell}_i^1, \hat{\ell}_i^2, \ldots, \hat{\ell}_i^{|\tilde{r}_i|}\big]$. 

After generating the student logits corresponding to all the examples in $\big[\mathcal{T}, \mathcal{S}\big]$, we formulate the loss as temperature-scaled KL divergence over the corresponding distributions over the entire sequence. We define:
\begin{align}
    \mathcal{L} = \frac{1}{|\mathcal{T}|}\sum_{i}T_{\text{KD}}^2\frac{1}{|\tilde{r}_i|}\sum_{j}\mathrm{KL}\left(\mathrm{softmax}\bigg(\frac{\ell_i^j}{T_{\text{KD}}}\bigg)\bigg|\bigg|\mathrm{softmax}\bigg(\frac{\hat{\ell}_i^j}{T_{\text{KD}}}\bigg)\right),
\end{align}
for a temperature $T_{\text{KD}} > 1$. No cross-entropy on hard labels is used. Thus, training is purely driven by matching the teacher's soft distributions over the answer tokens. Prompt tokens are never supervised directly. 

\begin{algorithm}[H]
    \caption{Semantic Soft Bootstrapping (SSB)}
    \label{algo: ssb}
    \begin{algorithmic}[1]
        \Require 
            Base LLM $f_\theta$; response $\sim f_\theta$(system\_prompt, query; temperature) \\
            Supervised dataset $\mathcal{D} = \{(q_i, a_i)\}$ with problems $q_i$ and ground-truth final answers $a_i$ \\
            Number of teacher rollouts $K$; rollout temperature $T_{\text{roll}}$, distillation temperature $T_{\text{KD}}$ \\
            $\text{sys}_{\mathrm{inf}}$ prompt instructs the LLM to generate the final answer within \verb|\boxed{...}| \\
            Learning rate $\alpha$; number of epochs = $E$
            
        \Statex
        \Statex \textbf{Phase 1: Logit Generation}
        \State \textbf{Teacher Reply Generation:}
        \For{each $(q, a) \in \mathcal{D}$}
            \State $\mathcal{R} \gets \emptyset$ \Comment{Store rollouts and parsed answers}
            \For{$k = 1, \dots, K$} \Comment{Teacher rollouts on question only}
                \State Sample teacher response $r_k \sim f_\theta(\text{sys}_{\mathrm{inf}}, q; T_{\text{roll}})$
                \State Parse $\hat{a}_k \gets \textsc{ExtractBoxedAnswer}(r_k)$
                \State $\mathcal{R} \gets \mathcal{R} \cup \{(r_k, \hat{a}_k)\}$
            \EndFor
            \State $\mathcal{R}^{\mathrm{correct}} \gets \{r_k \mid (r_k, \hat{a}_k) \in \mathcal{R},\ \hat{a}_k = a\}$
            \State $\mathcal{R}^{\mathrm{wrong}} \gets \{r_k \mid (r_k, \hat{a}_k) \in \mathcal{R},\ \hat{a}_k \neq a\}$
            \If{$\mathcal{R}^{\mathrm{correct}} = \emptyset$ \textbf{ or } $\mathcal{R}^{\mathrm{wrong}} = \emptyset$}
                \State \textbf{continue} \Comment{No mix of correct and incorrect responses; skip}
            \EndIf

            \State Choose one correct rollout $r^{\mathrm{corr}} \in \mathcal{R}^{\mathrm{correct}}$
            \State Let $\hat{a}^{\mathrm{wrong}}$ be the most frequent wrong answer in $\mathcal{R}^{\mathrm{wrong}}$ (tie broken at random)
            \State Choose $r^{\mathrm{wrong}} \in \mathcal{R}^{\mathrm{wrong}}$ with answer $\hat{a}^{\mathrm{wrong}}$
            \State SSB prompt $u_{\mathrm{SSB}}=$
            \big[problem $q$; correct trace $r^{\mathrm{corr}}$; incorrect trace $r^{\mathrm{wrong}}$; robust response?\big]
            \State Generate robust teacher response $\tilde{r} \sim f_\theta(\text{sys}_{\mathrm{inf}}, u_{\mathrm{SSB}}; T_{\text{roll}})$
            \State Parse $\tilde{a} \gets \textsc{ExtractBoxedAnswer}(\tilde{r})$
            \If{$\tilde{a} \neq a$}
                \State \textbf{continue} \Comment{Final robust answer is incorrect; discard}
            \EndIf

            \State Append teacher example:
            \Statex \qquad\qquad $\mathcal{T} \gets \mathcal{T} \cup \{\text{messages} = [(\text{system}: \text{sys}_{\mathrm{inf}}), (\text{user}: u_{\mathrm{SSB}}), (\text{assistant}: \tilde{r})]\}$
            \State Append student example:
            \Statex \qquad\qquad $\mathcal{S} \gets \mathcal{S} \cup \{\text{messages} = [(\text{system}: \text{sys}_{\mathrm{inf}}), (\text{user}: q), (\text{assistant}: \tilde{r})]\}$
        \EndFor
        \State \textbf{Teacher logit precomputation:}
        \For{each teacher example index $i$ with messages $m_i \in \mathcal{T}$}
            \State $m_i^{\mathrm{teacher}} \gets m_i$ without the final assistant message $\tilde{r}_i$
            \For{$j = 0, $\ldots$ , |\tilde{r}_i|-1$}
            \State logit $\ell_i^j \gets f_\theta(m_i[0:|m_i^{\mathrm{teacher}}|+j])$ \Comment{teacher logit generation}
            \EndFor
            \State Save $\ell_i = \big[\ell_i^1, \ell_i^2, \ldots, \ell_i^{|\tilde{r}_i|}\big]$ to disk
        \EndFor

        \Statex
        \Statex \textbf{Phase 2: Logit Distillation}
        \For{epoch $= 1, 2, \ldots, E$}
            \For{each student example index $i$ with messages $s_i \in \mathcal{S}$}
                \State $m_i^{\mathrm{student}} \gets s_i$ without the final assistant message $\tilde{r}_i$
                \For{$j = 0, $\ldots$ , |\tilde{r}_i|-1$}
                \State logit $\hat{\ell}_i^j \gets f_\theta(s_i[0:|m_i^{\mathrm{student}}|+j])$ \Comment{student logit generation}
                \EndFor
                \State $\hat{\ell}_i \gets \big[\hat{\ell}_i^1, \hat{\ell}_i^2, \ldots, \hat{\ell}_i^{|\tilde{r}_i|}\big]$
                \State Calculate the KD loss:
                \Statex \qquad\qquad\quad$\mathcal{L}_i \gets T_{\text{KD}}^2 \frac{1}{|\tilde{r}_i|}\sum_{j}\mathrm{KL}(\mathrm{softmax}(\ell_i^j/T_{\text{KD}})||\mathrm{softmax}(\hat{\ell}_i^j/T_{\text{KD}}))$ \Comment{logit matching}
            \EndFor
            \State $f_{\theta} \gets f_{\theta} - \alpha g\left(\frac{1}{|\mathcal{T}|}\nabla_\theta\sum_i\mathcal{L}_i\right)$ \Comment{$g(\cdot)$ depends on optimization algorithm}
        \EndFor
        \State \Return trained model $f_{\theta}$
    \end{algorithmic}
\end{algorithm}

This whole logit generation and distillation process is demonstrated in the Algorithm~\ref{algo: ssb}. The key insight here is that distillation arises purely from a change in semantic context, not from a larger or separate teacher model. The same model acts as both the teacher and the student; the only thing that differs is: the teacher sees the problem with explicit correct and incorrect solutions and is asked to synthesize a didactic, error-aware explanation. While the student sees only the original problem and is trained to emulate the teacher's logit-level behavior on the answer sequence. The model thus uses its own prior solutions (both successful and mistaken) to construct higher-quality teaching signals, and then distills those signals into a student that must perform well without access to those hints. In this way, SSB can be viewed as distilling richer internal semantics (``hints'') into the model's parameters, such that the improved behavior is available at inference time from the question alone.

\section{Experiments}
\subsection{Settings}
We implement SSB by utilizing parameter-efficient fine-tuning (PEFT) by \cite{unsloth} to bootstrap unsloth/Qwen-2.5–3B–Instruct base model. We use rank 32 LoRA to update around 2\% of total model parameters for the fine-tuning. We use GSM8K dataset, which is a collection of grade school level math questions and answers, for generating sample model rollouts. We do not use any of the sample answer responses available in the GSM8K dataset, only the question and the final answer. For each question, we generate four such samples and categorize them into two sets: correct and incorrect responses. After the successful robustification of the responses, we save the answers into teacher and student example sets. In this way, we curate 256 sample examples in the teacher–student format by processing 950 questions. We train the SSB algorithm with batch size of 4, and for 3 epochs, for a total of 192 steps. For the GRPO setting, which is used as a control, we use 2000 samples from GSM8K dataset. Both the SSB and GRPO trainings were performed using a single NVIDIA A100 40 GB GPU.

\subsection{Results}
\begin{table}[ht]
  \caption{Pass@1 accuracy comparisons in benchmarks}
  \label{table: results}
  \centering
  \begin{tabular}{lcc}
    \toprule
    \multirow{2}{*}{Model} & \multicolumn{2}{c}{Dataset} \\
    \cmidrule(lr){2-3}
           & MATH500 & AIME2024\\
    \midrule
    unsloth/Qwen2.5-3B-Instruct (base model) & 37.6\% & 0.0\%\\
    GRPO training & 44.8\% & 3.33\%\\
    \rowcolor{gray!20}
    SSB training & 55.4\% & 13.33\%\\
    \bottomrule
  \end{tabular}
\end{table}

We measured the accuracy of the base model, the SSB trained model, and the GRPO trained model on MATH500, and AIME2024 benchmarks. For calculating accuracy, we use the simple pass@1 format~\citep{chen2021evaluating}, which is an estimation of the probability of getting correct answer by the model in its first attempt. It is defined as: 
\begin{align}
    \text{Pass@1 accuracy} = \frac{\text{\# correct answers}}{\text{\# questions}} = \frac{1}{L}\sum_{j=1}^{L}\mathbb{I}_j, 
\end{align}
where $L$ is the total number of questions in the benchmark; $\mathbb{I}_j = 1$, if the boxed answer to the $j$th question is in the response from the model, and $0$, otherwise. The results of our experiment are shown in Table~\ref{table: results}. We observe that SSB training outperforms GRPO training by 10.6\% and 10\% on MATH500 and AIME2024 benchmarks, respectively.

\begin{figure}[t]
  \centering
  \begin{subcaptionbox}{Loss over training steps.\label{fig: ssb_loss}}[0.45\linewidth]
    {\includegraphics[width=\linewidth]{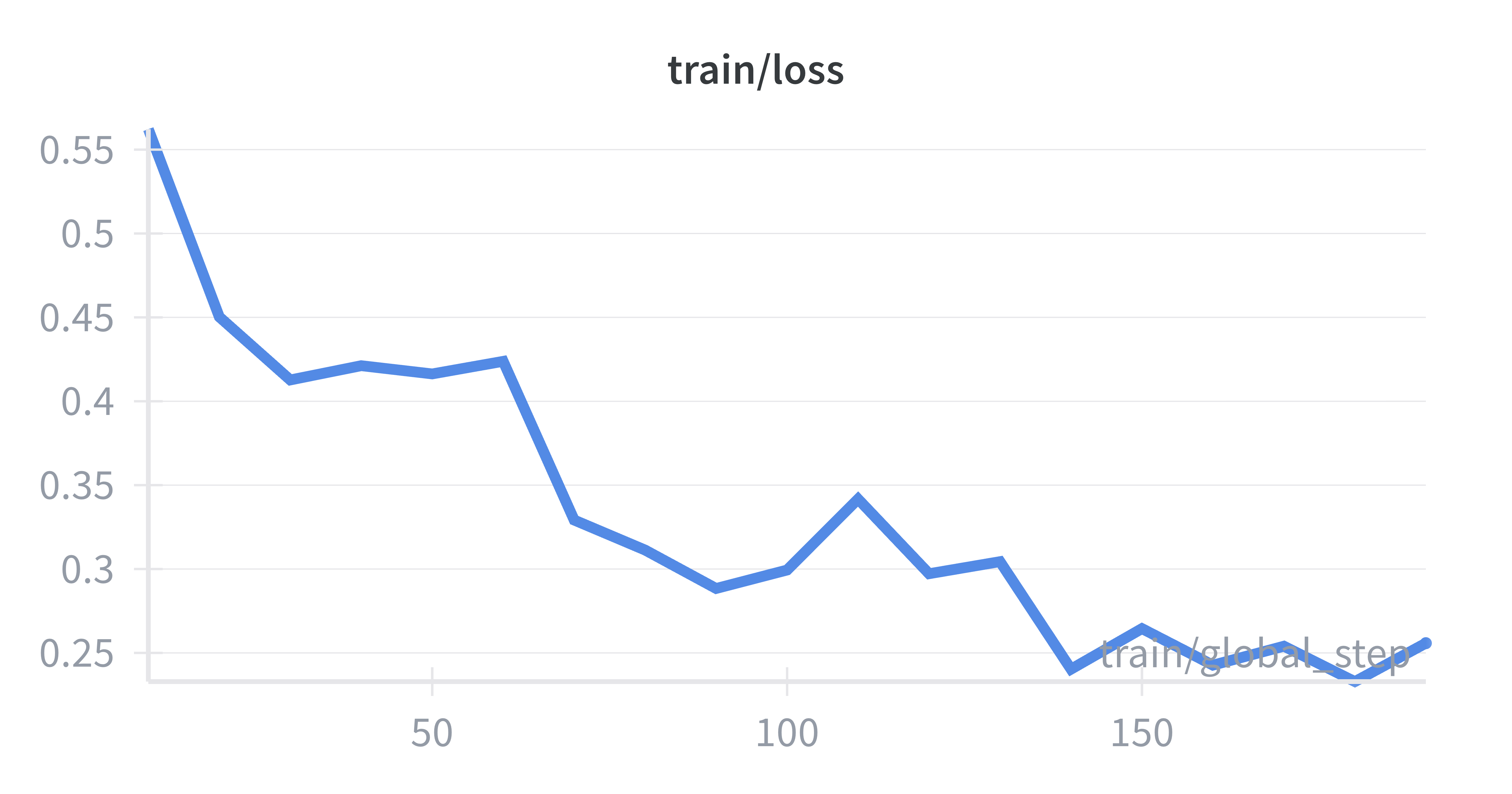}}
  \end{subcaptionbox}
  \hspace{0.05\linewidth}
  \begin{subcaptionbox}{Gradient norm over training steps.\label{fig: ssb_grad}}[0.45\linewidth]
    {\includegraphics[width=\linewidth]{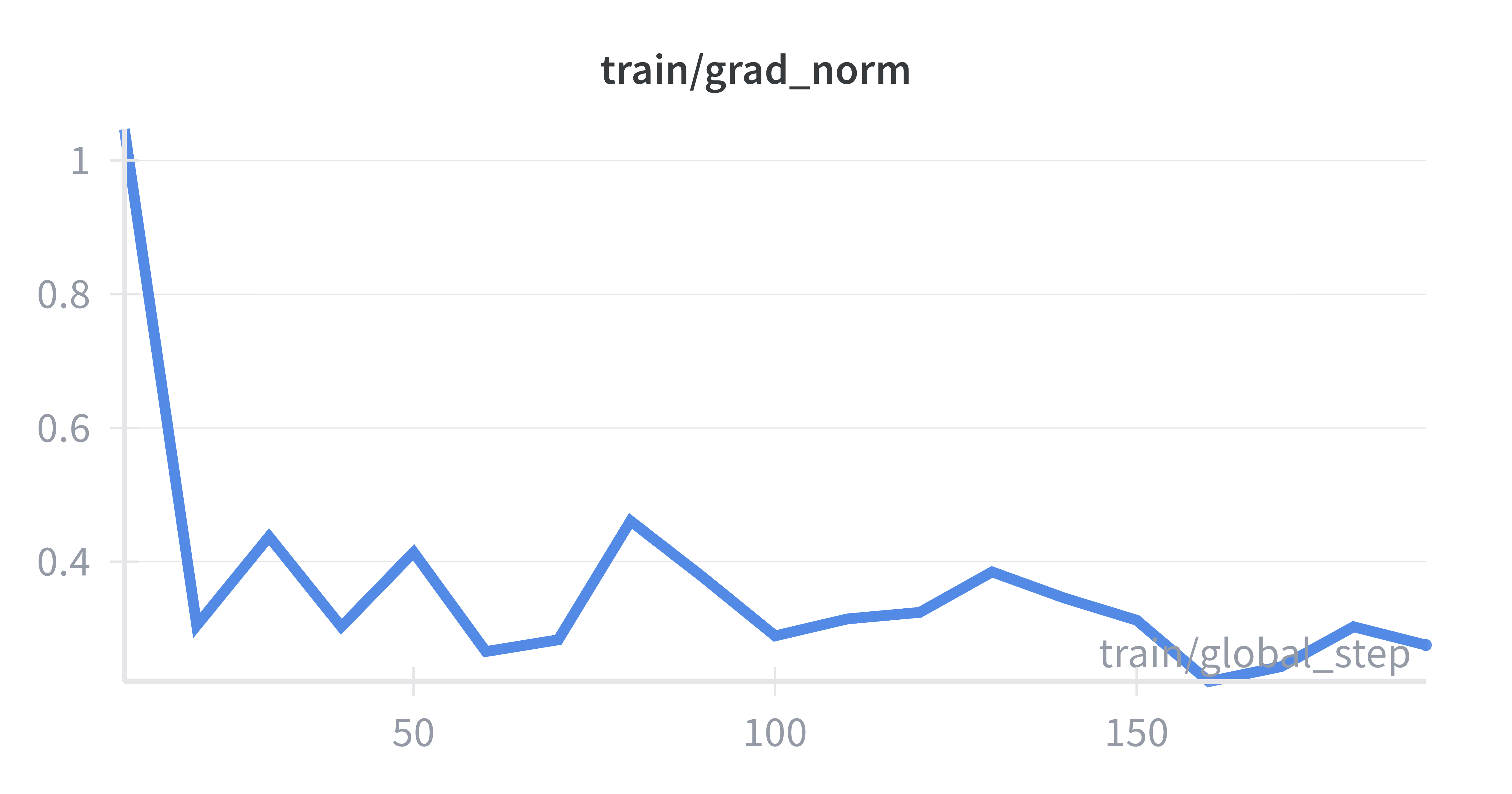}}
  \end{subcaptionbox}
  \hspace{0.05\linewidth}
  \begin{subcaptionbox}{Completion length over training steps.\label{fig: ssb_length}}[0.45\linewidth]
    {\includegraphics[width=\linewidth]{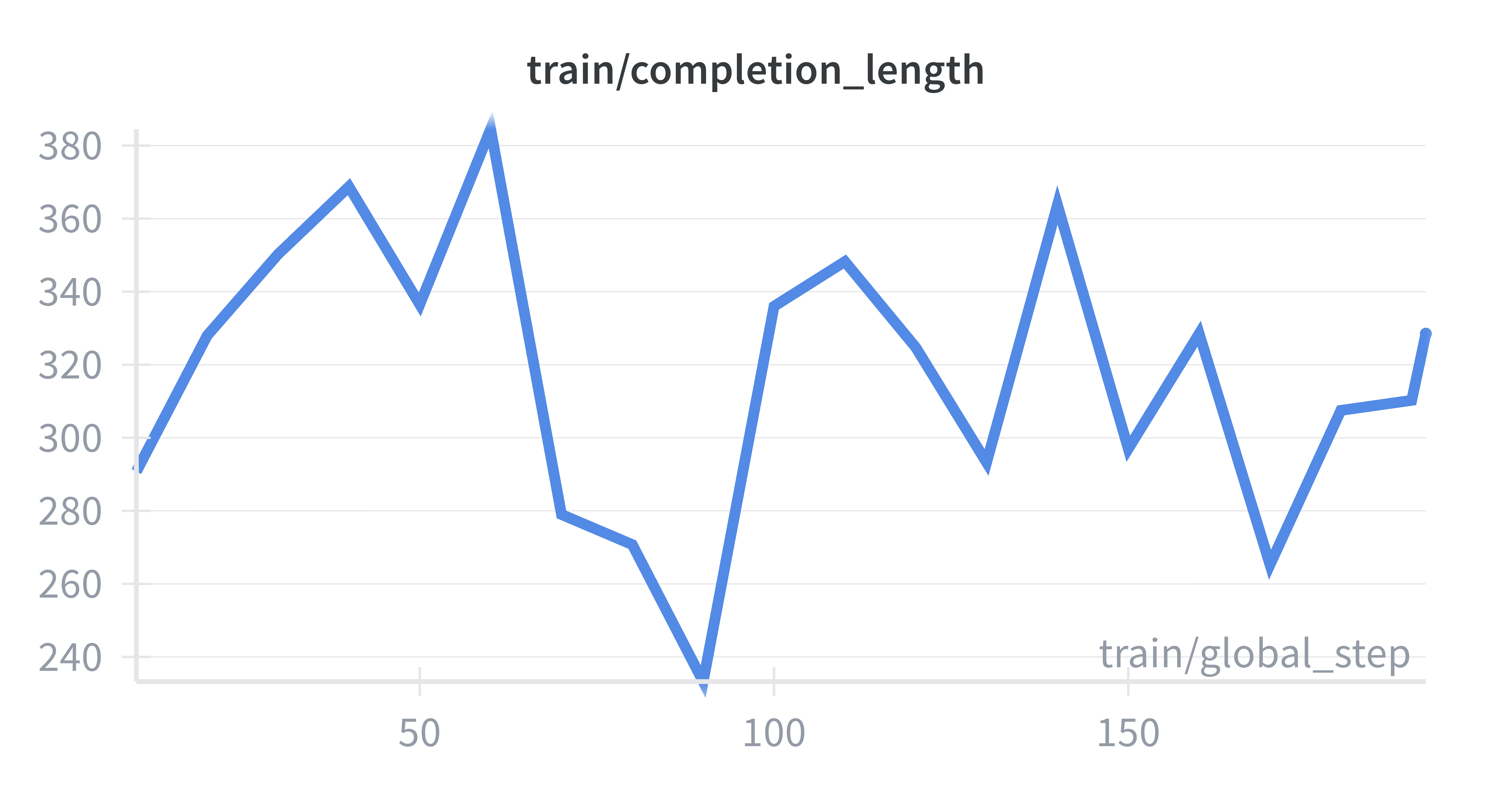}}
  \end{subcaptionbox}
  \caption{Training evolution in SSB shows that the model training is fast and stable; the overall completion length does not increase much.}
  \label{fig: ssb_train}
\end{figure}

Furthermore, we plot the training progression of the SSB training over 192 steps in Fig.~\ref{fig: ssb_train}. We observe in Fig.~\ref{fig: ssb_loss} that the loss decreases gradually as training step increases, indicating a stable training dynamics. A similar trend is observed for the gradient norm in Fig.~\ref{fig: ssb_grad}, which indicates convergence. However, for completion length in Fig.~\ref{fig: ssb_length}, we do not observe any significant increase over training steps. This is not what we usually observe in GRPO training, or any kind of RLVR training, in general. Combined with the performance in benchmarks, we conclude that increase in average response length, and thus, increased token usage, is not a necessary indicator of increase in reasoning capabilities.

\section{Conclusion}
In this work, we introduce \emph{Semantic Soft Bootstrapping} (SSB), an RL-free self-distillation
framework that improves long-context reasoning in LLMs by training the model on its
own hinted reasoning as a teacher. Rather than relying on a separate larger teacher or on-policy
gradient with sparse rewards, SSB uses the same base model in two semantic roles: a hinted teacher
that sees both correct and incorrect solutions and synthesizes a robust explanation, and a hint-free
student that learns to reproduce this behavior from the bare question alone. Starting from a raw
problem–answer dataset, we construct paired teacher–student conversations and then
precompute teacher logits over the answer tokens, enabling efficient offline distillation without any
human annotation or online RL loop. Our experiments with Qwen2.5-3B-Instruct on GSM8K, MATH500, and AIME2024 show that SSB achieves substantial gains in pass@1 accuracy over a GRPO baseline; on MATH500 and
AIME2024 by 10.6\% and 10\%, respectively, relative to GRPO. SSB exhibits stable training
dynamics and no systematic increase in completion length. These results suggest that stronger
reasoning does not require ever-longer chains-of-thought or token-intensive RLVR runs, and that
logit-level supervision on verified answer trajectories is a powerful and compute-efficient alternative
for post-training reasoning models. We believe this work can be scaled further to larger models with more compute power. Furthermore, it can be extended to a more diverse environment of broader domains, e.g., program synthesis and scientific questions and answers. It is insightful to study the sample efficiency and scaling laws of SSB as the number of model parameters, curated problems, domains and rollouts grow, and to compare its compute–accuracy tradeoffs more systematically against modern RLVR pipelines. 

\bibliography{references}  
\end{document}